\documentclass[10pt,twocolumn,letterpaper]{article}

\usepackage[pagenumbers]{cvpr} 

\usepackage{graphicx} 
\usepackage{blindtext}
\usepackage{graphicx}
\usepackage{algorithm2e}
\RestyleAlgo{ruled}
\usepackage{fullpage}
\usepackage{latexsym,amsmath,amssymb}    
\usepackage{newfloat}
\usepackage{listings}
\usepackage{enumitem}
\usepackage{subcaption}
\usepackage{booktabs}[demo]

\usepackage[pagebackref,breaklinks,colorlinks]{hyperref}

\usepackage[capitalize]{cleveref}
\crefname{section}{Sec.}{Secs.}
\Crefname{section}{Section}{Sections}
\Crefname{table}{Table}{Tables}
\crefname{table}{Tab.}{Tabs.}

\def\cvprPaperID{*****} 
\def\confName{CVPR}
\def\confYear{2023}

\begin{document}

\title{A Practical Mixed Precision Algorithm for Post-Training Quantization}

\author{Nilesh Prasad Pandey, Markus Nagel, Mart van Baalen, \\Yin Huang, Chirag Patel, Tijmen Blankevoort \\
Qualcomm AI Research$^{*}$\\
{\tt\small \{nileshpr, markusn, mart, yinh, cpatel, tijmen\}@qti.qualcomm.com}}

\maketitle
\begin{abstract}
Neural network quantization is frequently used to optimize model size, latency and power consumption for on-device deployment of neural networks. In many cases, a target bit-width is set for an entire network, meaning every layer get quantized to the same number of bits. However, for many networks some layers are significantly more robust to quantization noise than others, leaving an important axis of improvement unused. As many hardware solutions provide multiple different bit-width settings, mixed-precision quantization has emerged as a promising solution to find a better performance-efficiency trade-off than homogeneous quantization. However, most existing mixed precision algorithms are rather difficult to use for practitioners as they require access to the training data, have many hyper-parameters to tune or even depend on end-to-end retraining of the entire model. In this work, we present a simple post-training mixed precision algorithm that only requires a small unlabeled calibration dataset to automatically select suitable bit-widths for each layer for desirable on-device performance. Our algorithm requires no hyper-parameter tuning, is robust to data variation and takes into account practical hardware deployment constraints making it a great candidate for practical use. We experimentally validate our proposed method on several computer vision tasks, natural language processing tasks and many different networks, and show that we can find mixed precision networks that provide a better trade-off between accuracy and efficiency than their homogeneous bit-width equivalents.
\end{abstract}
\let\thefootnote\relax\footnote{*Qualcomm AI Research is an initiative of Qualcomm Technologies, Inc}

\section{Introduction}
Due to the ever increasing computational and memory cost of the networks, methods such as quantization, pruning and efficient network design have gained considerable attention in the literature to compress and facilitate deployment of these models on low computational resource devices. In this paper, we focus specifically on quantization, where the parameters and operations are kept and conducted in lower bit-widths than the 32 bit formats generally used for neural network training \cite{nagel2021white}. This process is not free, as quantizing parameters and operations leads to noise being introduced in the network, which in turn can lead to a degradation in performance of the network itself. The problem of quantization is thus to find a network that runs faster in practice, while retaining as much of the accuracy lost due to the introduction of quantization noise. 


A common practice is to quantize all layers in a network to the same bit-width. However, for many networks there is a distinct difference in sensitivity to quantization from layer to layer. Mixed precision quantization aims at solving the quantization issue by keeping more sensitive layers in higher precision while maintaining the rest of the network in lower bits, effectively improving the performance-efficiency trade-off of the network. 

Most approaches in this domain~\cite{yang2021fracbits,huang2022sdq,dong2019hawq} start with a pretrained network, learn the bit-width assignments and finetune the quantized network to get the final mixed precision model. However, this process requires access to task training data and the resources to run extensive training and hyperparameter tuning to get the final mixed precision model. These methods also consider all weight layers and activations to be independent and search mixed precision policies for a network without considering the constraints for on device deployment making it a non-favorable choice for practical post-training scenarios.

In this work, we introduce a post-training quantization algorithm that sets the mixed-precision bit-widths for real-world practical use-cases. It uses little data, works efficiently, and takes into account practical hardware considerations. The end-result is a very practical algorithm that is highly effective in practice.

\vspace{-0.10in}

\section{Related Work}
Many methods have been proposed in the model efficiency space to solve the problem of neural network quantization~\cite{nagel2021white, banner2019post, nagel2019data, krishnamoorthi2018quantizing,esser2019learned}, fixed precision or homogeneous quantization uses same bit-widths for all layers in the network and can be categorized as post-training quantization~\cite{nagel2019data,banner2019post} or quantization-aware-training based approaches~\cite{esser2019learned,jacob2018quantization,nagel2022overcoming,bhalgat2020lsq+}. The quantization-aware training methods train a network with simulated quantization in-the-loop, to optimize them for quantized inference. The focus of this paper is post-training quantization, where the network is optimized for quantization without any re-training, and with less data and compute available. 

Most mixed precision quantization methods belong to either search based or optimization based approaches. Methods such as HAWQ \cite{dong2019hawq}, OMPQ~\cite{ma2021ompq} compute sensitivity metrics based on layers' hessian spectrum and orthogonality, respectively, to determine the bit-width configurations. Instead, we choose to directly measure the sensitivity of each layer on a network by evaluation, which is efficient, and is exact as it does not rely on estimates. 
Other methods formulate mixed precision quantization as a optimization problem by tackling the non-differentiability of bit-widths~\cite{huang2022sdq,yang2021fracbits}. Approaches like DNAS\cite{wu2018mixed}, HAQ~\cite{wang2019haq} use reinforcement learning and incorporate hardware feedback to solve this optimization problem. On the other hand, work such as~\cite{przewlocka2022power} use non-uniform quantization schemes to improve representational capability for lower bit-widths, but implementing such schemes is often not hardware friendly. Most of these approaches require quantization-aware-training to achieve the mixed precision model which makes these methods compute heavy and also time consuming. Our proposed work falls in the category of post-training quantization and search based uniform quantization.

\section{Method}
In this section we introduce our two-phase algorithm to solve the problem of mixed precision quantization in the post-training quantization setting. In the first phase, we create a per-layer sensitivity list by measuring the loss of the entire network with different quantization options for each layer. This list gives an estimate of the impact of quantizing for a layer on the network's performance. The second phase of the algorithm starts with the entire network quantized to the highest possible bit-width, after which based on the sensitivity list created in phase 1, we iteratively flip the least sensitive quantizers to lower bit-width options untill the performance budget is met or our accuracy requirement gets violated. 

As discussed in the later sections, labels have no role in creating the per-layer sensitivity list obtained in phase 1 of our method, making the algorithm agnostic to the category of the data used for calibration. Also, due to the simple iterative nature of phase 2 of our algorithm, our method seamlessly allows a user to pick either an on-device accuracy target or a performance budget by incorporating practical hardware deployment constraints in the search.

\subsection{Preliminaries}
In neural network quantization, real valued weight $W^{r}$ and activation tensors $x^{r}$ are quantized to an appropriate low precision value. The quantization operation for a $b$-bit uniform quantizer $q_{b}$ is defined as
\begin{align}
    W_{int} &= Clip \left(  \Big \lfloor{\frac{W^{r}}{s}}\Big \rceil, n, p \right) \\
    W^{r} &\approx q_{b}(W^{r}) = sW_{int} \hfill
\end{align}

where $s$ denotes the quantization scale parameter and, $n$ and $p$ denote the negative and positive integer thresholds for clipping. The same principle holds for per-channel scheme of quantization \cite{krishnamoorthi2018quantizing}, for which the scale factor $s$ is a vector with each value representing scale of each individual channel. For a more detailed coverage, please refer to \cite{nagel2021white}.

\subsection{Phase 1: Generating Sensitivity List}
In order to perform mixed precision quantization on a pretrained neural network, it is important to understand the effect of quantizing a layer on the network's performance. The first phase of our algorithm aims at creating such a per-layer sensitivity list. This list captures the relative sensitivity of layers to quantization, without taking into account any correlations with other layers for efficiency.

Next to the accuracy of the estimate of this list, for practical use-cases there are two more important factors for the creation of this list; the run-time of the algorithm, and the amount of data necessary to make a good estimate. In this section we propose to use the signal to quantization noise ratio (SQNR) as a metric to measure the relative sensitivity of each quantizer $q$ when quantized to different bit-width options $b \in B$. For a converged pre-trained network, quantizing the network introduces noise at the output which increases the network loss and directly affects the logits and the task performance. Hence, in order to measure the sensitivity to quantization of each quantizer $q$ in the network,  we define,
\begin{equation}\label{e:SQNR} 
    \Omega_{q,b}^{SQNR} = 10\log \frac{1}{N}\sum_{i=1}^{N}\frac{\mathbb{E}[{F}_{\theta} (x_{i})^{2}]}{\mathbb{E}[e(x_{i})^{2}]}
\end{equation}
where, $\Omega_{q,b}^{SQNR}$ is the average SQNR at the output of the network using N calibration data points, and the quantization error
\begin{equation}
   e(x) = \mathcal{F}_{\theta} (x) - \mathcal{Q}_{q,b}(\mathcal{F}_{\theta} (x))
\end{equation}
  where $\mathcal{F}_{\theta}$ represents the full precision network and $\mathcal{Q}_{q,b}(\mathcal{F}_{\theta})$ represents the quantized network with the quantizer $q$ set to bit-width $b$ and rest of the layers and activations in full precision.
  
\subsection{Phase 2: Finding Mixed Precision Configuration}
Once we have the relative sensitivity of layers to quantization, the next challenge is to allocate bit-widths to each layer in the network for the desired performance budget. To solve this optimization problem in an efficient manner, we propose to use an iterative pareto frontier based approach in the phase 2 of our algorithm to obtain a network with best performance-efficiency trade-off meeting the performance criteria. 

Phase 2 of our algorithm starts with the entire network quantized to a baseline highest bit-width that gives the best task performance upon quantization. At each step, in-order to minimize the loss we incur due to reducing the precision of a quantizer to a lower bit-width, we use the sensitivity list to get the best candidate for minimizing this performance loss vs efficiency gain trade-off. This pareto frontier based approach allows us to search for the most efficient mixed precision configuration meeting the performance budget. The approach is greedy, but works very well in practice.

\begin{algorithm}[!ht]
    \small
  \textbf{Input}: Network with full precision quantizer $\left\{q_{l} \right \} _{l=1}^{L}$, bit-width candidates for mixed precision $\left\{b_{i}\right \} _{i=1}^{N}$, Performance budget $\gamma$ and evaluation criteria $\mathcal{E}$
  \\
  
  \textbf{Output}: Network with quantized quantizer $\left\{q_{l}^{Q} \right \} _{l=1}^{L}$ and corresponding bit-width allocations $\left\{b^{l}\right \} _{l=1}^{L}$
  \\
  
  
  \textbf{Phase 1}: 
  \\
  Initialize Sensitivity list $S$
  \\
  
  \For{quantizer q $\in \left\{q_{i}\right\}_{i=1}^{L}$}
    {
    \For{bit-width b $\in B - \left\{b_{baseline}\right\}$ }
        {
        Calculate $\Omega_{q,b}^{SQNR}$  \\
        Add $(q, b, \Omega_{q,b}^{SQNR})$ to $S$
        }
    }
  Sort sensitivity list $S$ (highest to lowest $\Omega_{q,b}^{SQNR}$)
  \\
  
  \textbf{Phase 2}: 
  \\
  Initialize network to baseline candidate\\
  Performance $\mathcal{P} = \mathcal{E}(network)$
  \\
  
  \For{$q,b \in S$}
  {
    
    Quantize $q$ to bit-width $b$\\
    Update $\mathcal{P} = \mathcal{E}(network)$
    
  \If{$\mathcal{P} < \gamma$}
    {
        return previous model
    }
  }
  
\caption{Post-Training Mixed Precision Algorithm}
\end{algorithm}

We define two budget criteria relevant to on-device deployment of neural networks that can be easily used with our mixed precision routine.
\subsubsection{Efficiency based budget:} Quantizing a network to lower bit-widths improves the latency and power consumption, hence improving the overall efficiency of the model when deployed on-device. In-order to measure efficiency of a mixed precision network independent of a specific target platform, we use Bit Operations (BOPs)~\cite{van2020bayesian} as a surrogate measure. BOPs are defined as:
\begin{equation}
   BOPS(\phi) = \sum_{ops_{i} \in network} bits(\phi_{i})MAC(op_{i})
\end{equation}
where $op_{i}$ represents operations in the network, $bits(\phi_{i})$ the bit-width associated with weights and activations for operation $op_{i}$ and $MAC(op_{i})$ represents the total of Multiply-Accumulate (MAC) operations for operation $op_{i}$. As discussed in \cite{van2022simulated}, Bits Operations (BOPs) correlates strongly with relative power consumption and hence a budget on BOPs can be used as a criteria to obtain a desirable efficiency target.

\subsubsection{Task Performance based budget:} For practical use-cases, achieving certain task performance can be used as a criteria to obtain a mixed precision network. A user can define the lowest performance of the network they would tolerate, and find the best bit-width setting to get the best on-device performance while allowing no more than their maximal tolerated degradation in performance. 

\subsection{Quantizer Groups in networks}
As described in the previous section, current methods in the mixed precision literature consider all weight layers and activations to be independent. This is generally not the case in practice. Not all options for quantizing operations in the network might be available. For example, a device might have only implemented kernels for W4A8, 4 bit weights and 8 bit activations, and W8A16, 8 bit weights and 16 bit activations. This creates a dependency for the specific kernels that can be picked by the mixed-precision algorithm. In this example, 4 bit weights always come with 8 bit activations, and 8 bit weights come with 16 bit activations. And other solutions are not valid.
To incorporate these practical constraints in our algorithm, we introduce the concept of a Quantizer Group which are groups of weights and activations connected through shared operations in a computational graph. To ensure these shared operations in a group are performed in a certain bit-width on-device, we constraint inputs to all the operation to be quantized to the same precision.

Incorporating the concept of quantizer groups in the current search based mixed precision \cite{huang2022sdq, yang2021fracbits} and metric based mixed precision methods \cite{dong2019hawq, ma2021ompq} can be non-trivial due to the design and computational complexity of the algorithms. Due to the iterative nature of our method, we can easily incorporate this practical constraint in our algorithm by measuring the per-group sensitivity and flipping the entire quantizer group instead of individual quantizers at each step of our phase 2.

\subsection{Improving post-training Mixed Precision for Low Bit Quantization with AdaRound}
For low bit ($<$8) quantization, it is very important to correctly capture the weight-activation quantization trade-off to obtain an optimal mixed precision configuration. AdaRound \cite{nagel2020up}, a post-training quantization algorithm which learns better rounding for weight quantization, has proven to be very effective in practice to improve low bit weight quantization performance of neural networks. Hence, in-order to improve the performance of our mixed precision method, we propose to integrate AdaRound into our algorithm.

To incorporate AdaRound, we propose to use AdaRounded weights to create the per-layer sensitivity of layers in Phase 1 of our algorithm. This allows us to obtain a sensitivity list with AdaRound and captures the correct trade-off between weight and activation quantization.
Due to the sequential and layer-wise optimization of the AdaRound algorithm, we can reuse the AdaRounded weights from Phase 1 and stitch the per-layer rounded weights together for each bit-width configuration we test in Phase 2. 
This allows to add AdaRound seamlessly into our mixed precision algorithm with minimal compute overhead. We will show this is very effective in improving the mixed precision network's performance.

\begin{table*}[ht]
    \centering
    \fontsize{8.25pt}{9.25pt}\selectfont
    \begin{tabular} {  lccccc }
    \toprule
    Model & FP32 & W8A8 &  PTQ MP &  W6A8  & PTQ MP\\
    &   &  ($r$=0.50) &  ($r$=0.50) &   ($r$=0.375)  &  ($r$=0.375)\\

    \midrule
    Resnet18 & $69.75\%$ & $69.56\%$ & $69.56\%$ & $69.27\%$ & $67.39\%$\\
    Resnet50 & $76.13\%$ & $75.95\%$ & $75.95\%$ & $75.41\%$ & $74.98\%$\\
    Mobilenetv2 & $71.87\%$ & $70.44\%$ & $70.68\%$ & $68.14\%$ &  $67.38\%$\\
    Mobilenetv3 & $74.04\%$ & $68.75\%$ & $71.65\%$ &$65.20\%$ &  $66.70\%$ \\
    Efficientnet-lite & $75.44\%$ &  $75.13\%$ & $75.14\%$ & $73.43\%$ & $73.64\%$\\
    Efficientnet-b0 & $77.67\%$ &  $12.40\%$ & $74.28\%$ & $12.62\%$ & $61.30\%$\\
    Deeplabv3-mobilenetv3 & $0.6887$ &  $0.5784$ & $0.6700$ & $0.5350$ & $0.6690$ \\
    BERT (MNLI) & $84.40\%$ &  $74.13\%$ & $82.97\%$ & $75.65\%$ & $82.22\%$\\
    ViT & $81.31\%$ &  $18.83\%$ & $80.58\%$ & $16.37\%$ & $77.08\%$\\

    \bottomrule
\end{tabular}
    \caption{MP using W4A8, W8A8, W8A16 bit-width candidates. Comparison between fixed precision quantization and mixed precision quantization.}
    \label{tab:fixed-vs-mixed}
\end{table*}

\begin{table*}
    \centering
    \fontsize{8.25pt}{9.25pt}\selectfont
    \begin{tabular} {  lccccc }
    \toprule
    Model & FP32 & W6A6  & PTQ MP &  W4A8   &  PTQ MP \\
          &   &  ($r$=0.281) &  ($r$=0.281) &   ($r$=0.25)  &  ($r$=0.25)\\
    \midrule
    Resnet18 & $69.75\%$ & $66.09\%$ & $65.14\%$ & $54.93\%$ & $63.14\%$ \\
    Resnet50 & $76.13\%$ & $70.54\%$ & $73.11\%$ & $66.82\%$ & $73.05\%$ \\
    Efficientnet-lite & $75.44\%$ &  $72.87\%$ & $74.17\%$ & $6.37\%$ & $73.49\%$\\
    Mobilenetv2 & $71.87\%$ & $58.60\%$ & $64.97\%$ & $9.55\%$ & $61.81\%$ \\
    Mobilenetv3 & $74.04\%$ & $4.19\%$ & $42.57\%$  & $2.83\%$ & $28.67\%$ \\

    \bottomrule
\end{tabular}
    \caption{MP using expanded search space: W4A4, W4A6, W6A4, W6A6, W8A6, W6A8, W8A8, W8A16 bit-width candidates. Comparison between fixed precision quantization and mixed precision quantization for low bit-widths.}
    \label{tab:fixed-vs-mixed_expanded}
\end{table*}

\subsection{Improving Phase 2 mixed precision configuration search}
Due to the sequential nature of our phase 2 algorithm, the overall run-time complexity to search for a desired mixed precision configuration is $\mathcal{O}(LM)$, where $L$ is the depth of the network and $M$ is the number of bit-width options for mixed precision. In-order to improve the run-time of our search algorithm, we propose to use  improved binary search based scheme exploiting the monotonic nature of the pareto curve obtained in phase 2.

\begin{figure}
    \centering
    \includegraphics[width=0.34\textwidth]{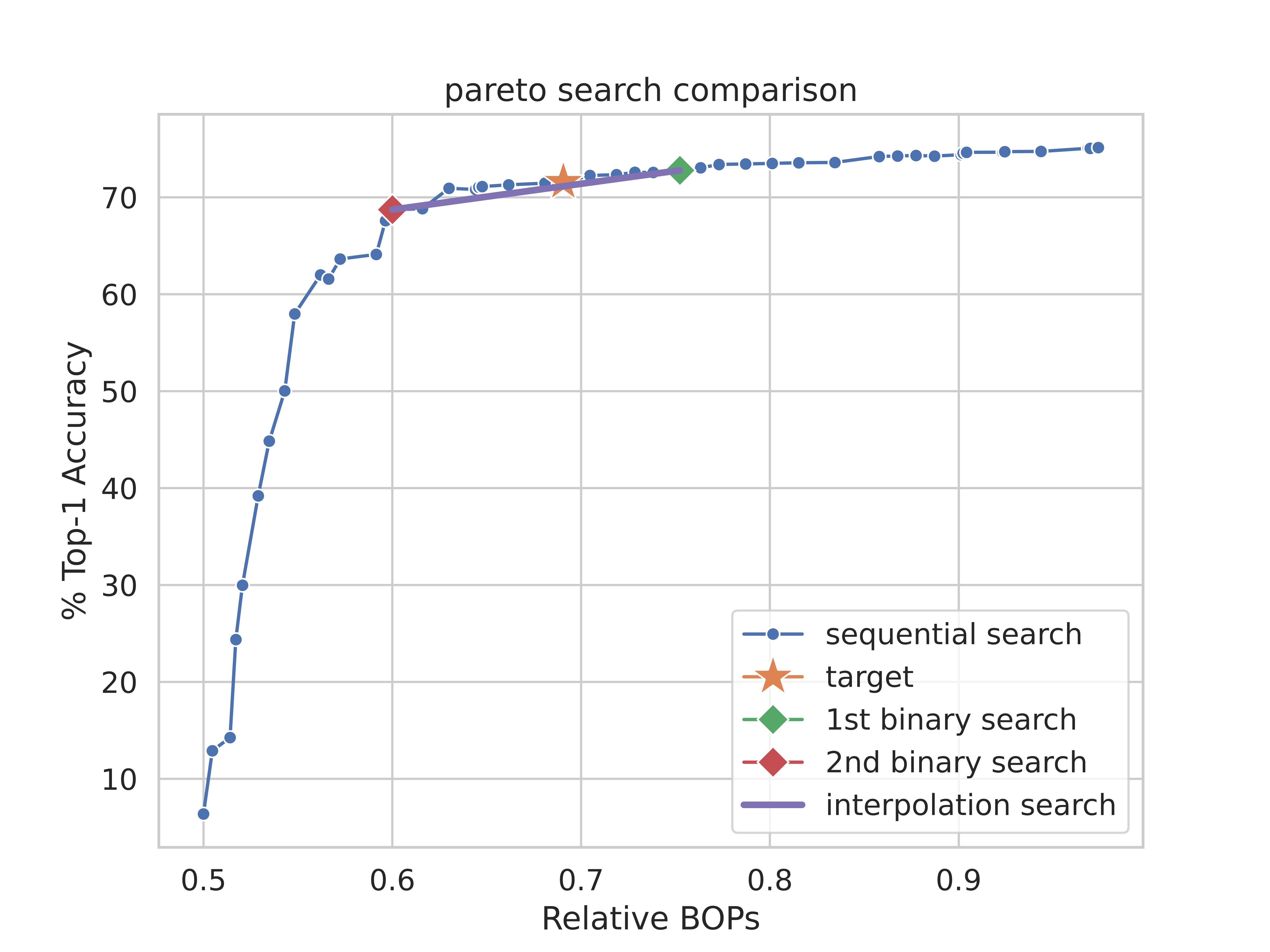}
    \caption{ Illustration of Binary + Interpolation Search.}
    \label{fig:hybridsearch}
\end{figure}

We start with a basic binary search  \cite{knuth1997art}, which reduces the overall search run-time complexity to $\mathcal{O}\left(\log_2(LM)\right)$. We further consider given the nonlinear nature of the pareto curve, first use base binary search to divide the entire curve into several piece-wise linear sub curves. For the linear sub curve, an interpolation search \cite{peterson1957addressing} can be employed to further reduce the search candidate to 1. In implementation, we use 2 times of binary search to divide the entire $LM$-points pareto curve to 4 $\lceil LM/4\rceil$-points sub curve candidates, then do the interpolation search on one of the candidate. Our improved hybrid search is illustrated in Figure \ref{fig:hybridsearch}.

\vspace{-0.10in}

\section{Experiments and Results}
\vspace{-0.08in}

In this section we evaluate the performance of our method on various models in different mixed precision settings. We start with defining our experimental setup by describing the datasets and the networks used, compare our mixed precision results to fixed precision quantization results and finally discuss various ablations to understand the benefits of our algorithm. 

\paragraph{Datasets:} For our experiments, we use the Imagenet-1K \cite{russakovsky2015imagenet}, Pascal VOC dataset \cite{everingham2015pascal} and GLUE benchmark \cite{wang2018glue}. Due to post-training nature of the algorithm, no training or fine-tuning is involved. Standard dataset based inference time augmentation pre-processing is used on the images.

\begin{figure*}
    \centering
    \subfloat[]{\includegraphics[width=0.25\textwidth]{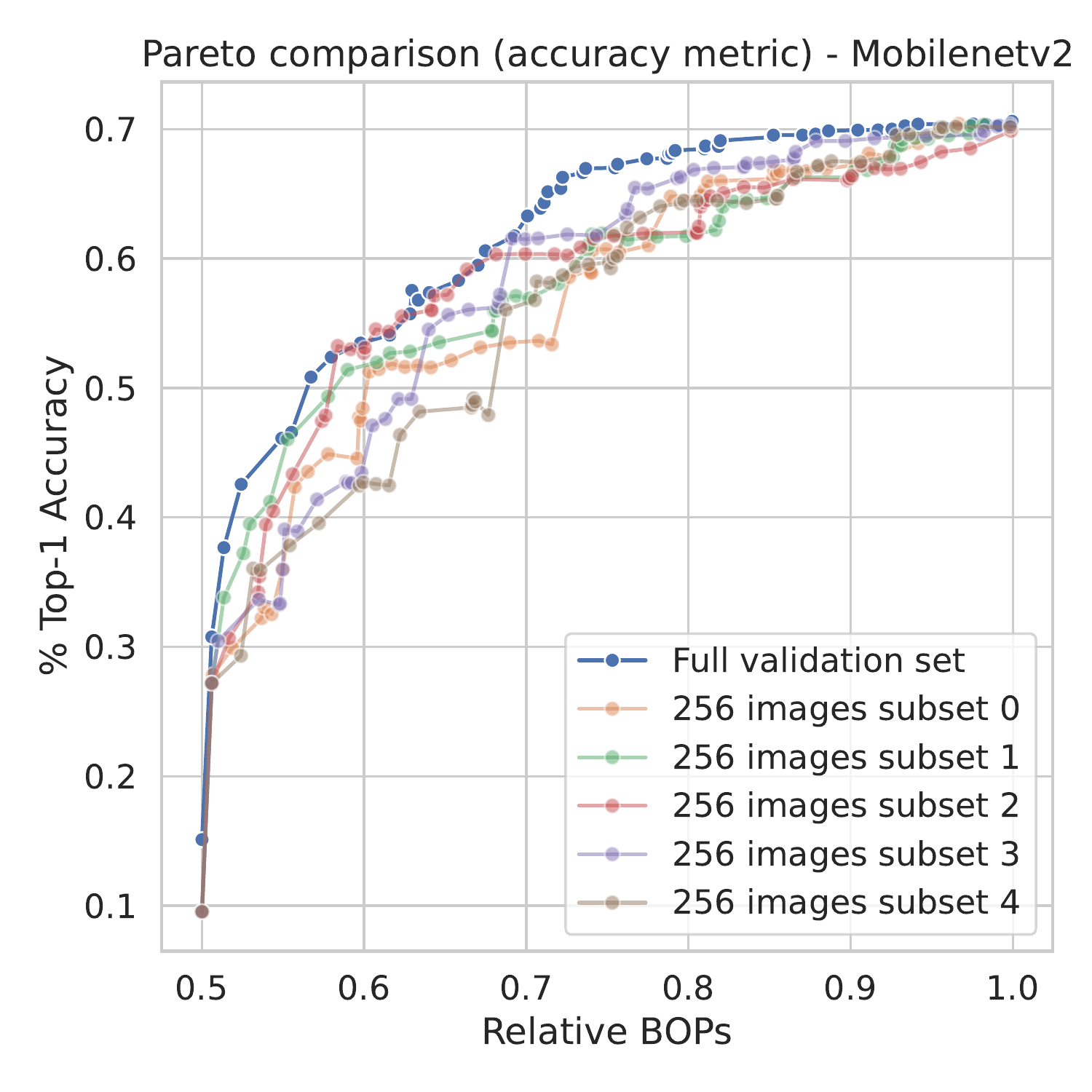}} 
    \subfloat[]{\includegraphics[width=0.25\textwidth]{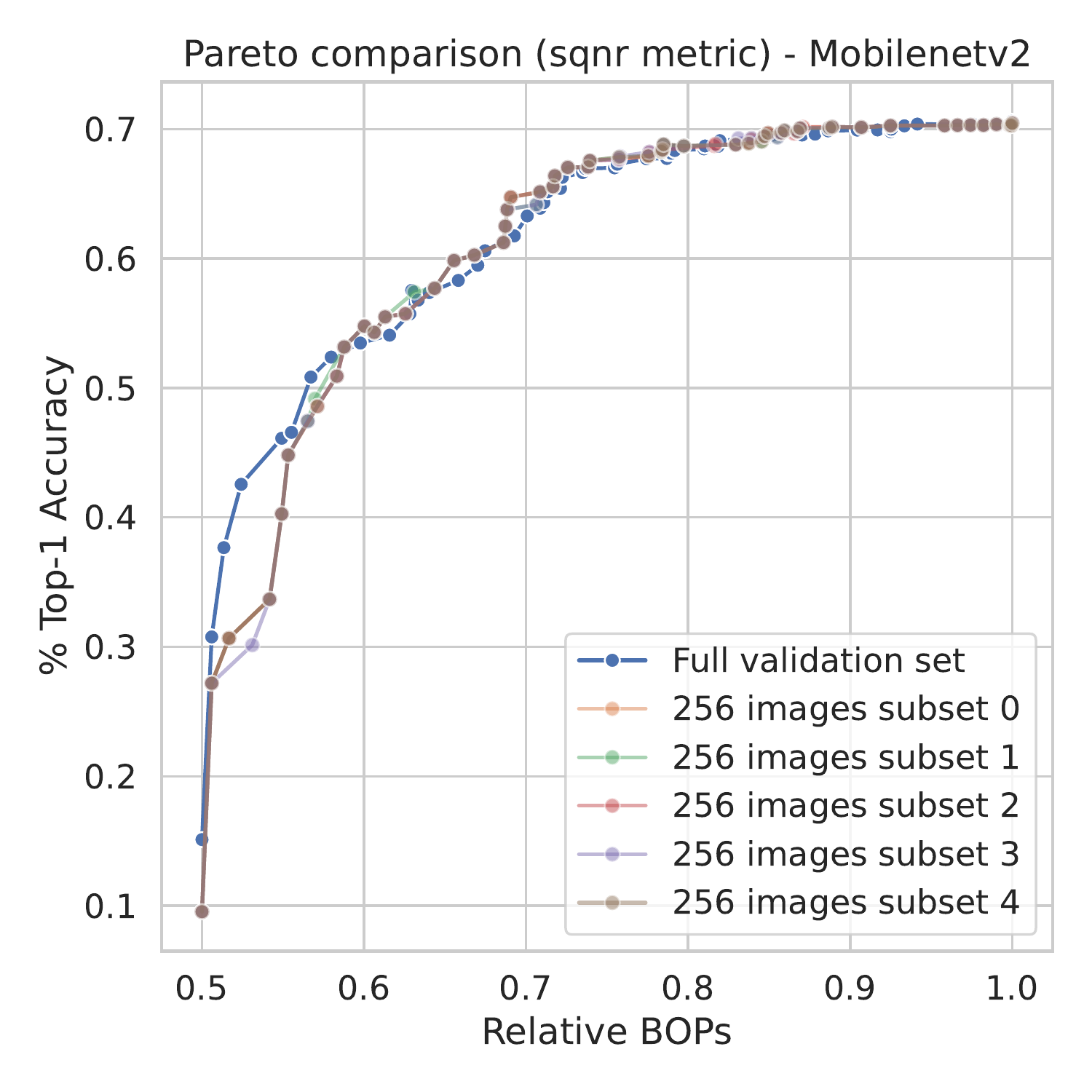}} 
    \subfloat[]{\includegraphics[width=0.25\textwidth]{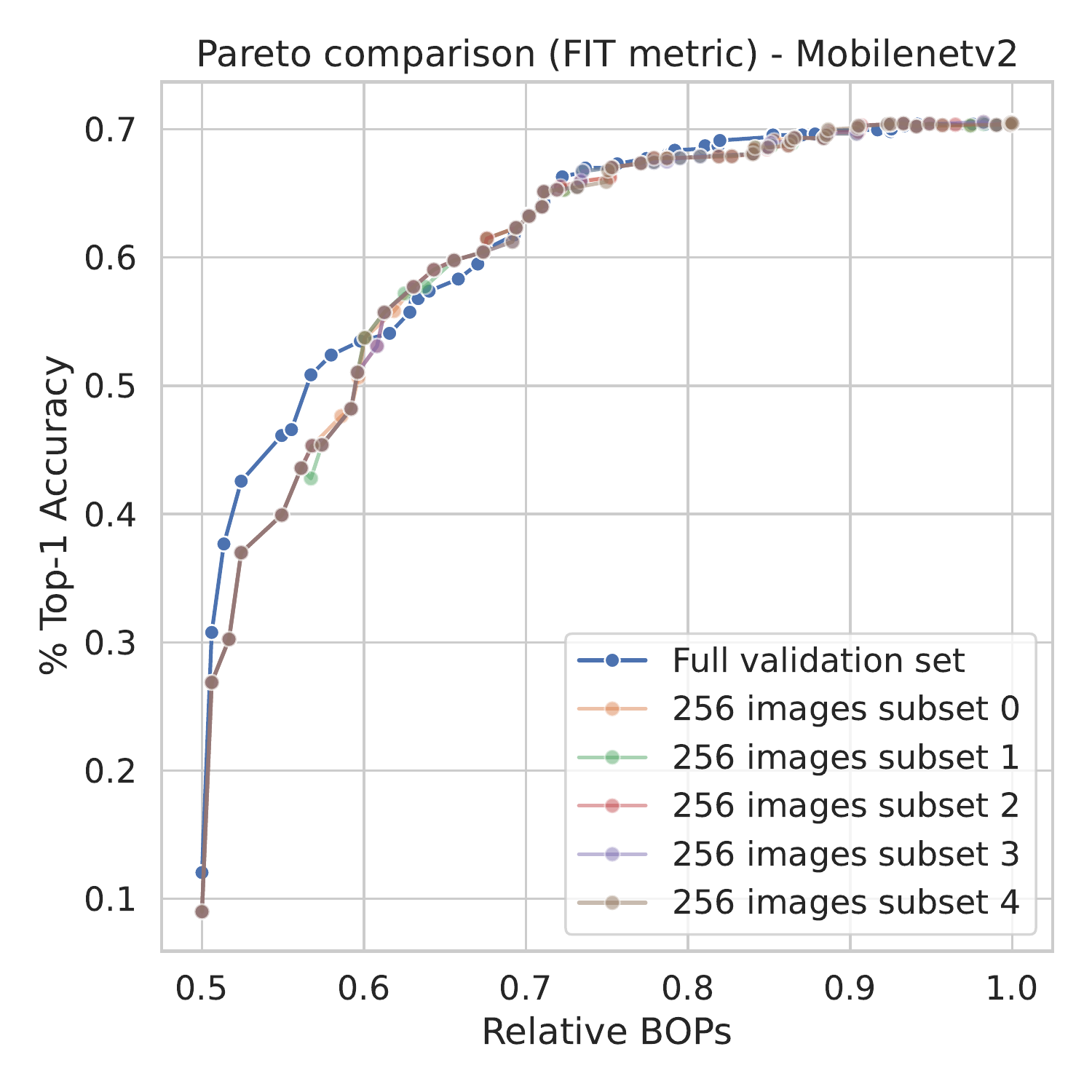}} 
    \subfloat[]{\includegraphics[width=0.25\textwidth]{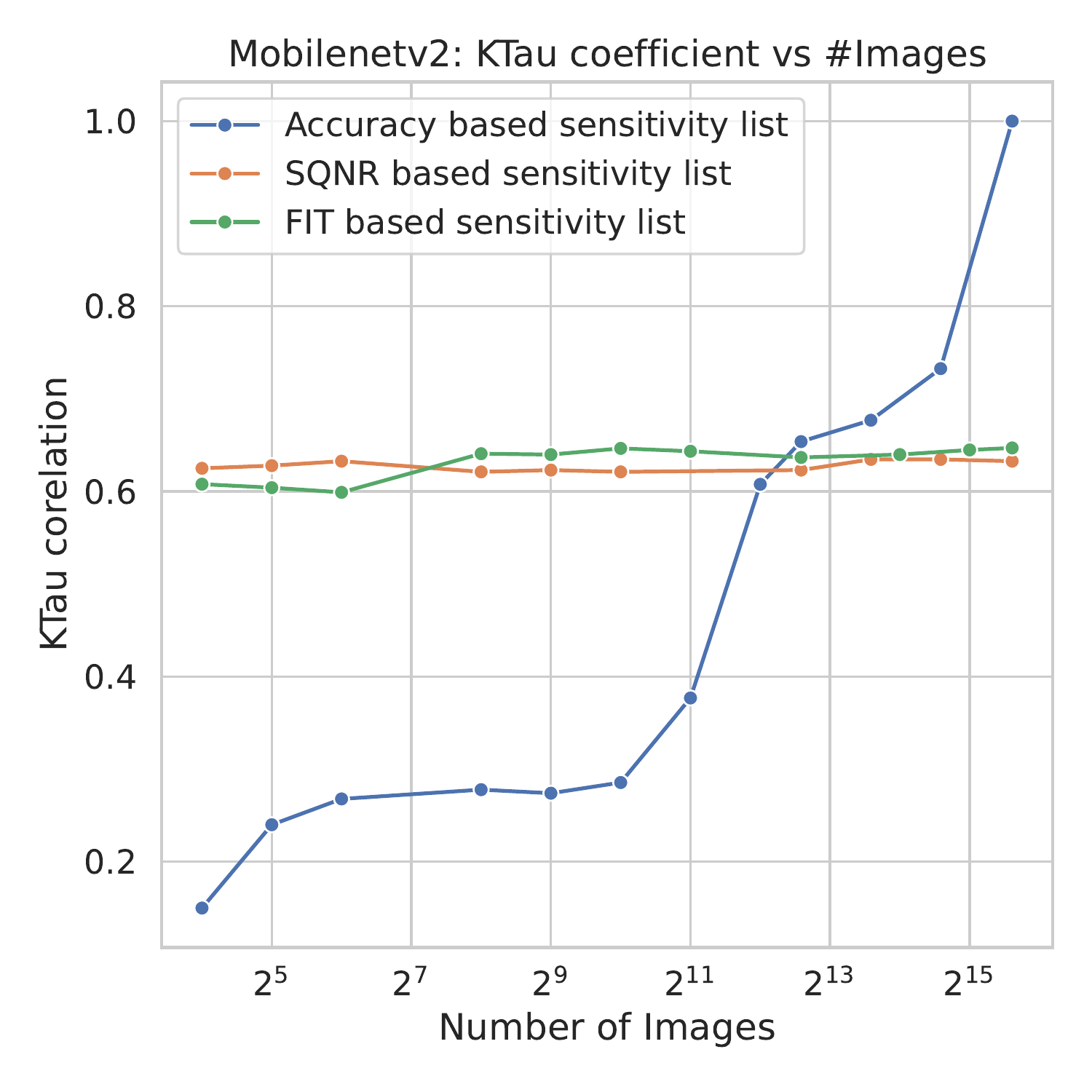}} 
    

    \caption{Variation in pareto curves obtained for MP using W4A8, W8A8 bit-width candidates using different subsets of 256 Imagenet images using (a) Accuracy (b) SQNR (c) FIT metric in Phase 1 to measure per-layer quantization sensitivity.  Compression achieved for obtained MP models is measured in terms of relative BOPs with respect to W8A8 model. (d) Kendall-Tau co-relation coefficient ($\tau$) of the sensitivity list obtained using accuracy, SQNR and FIT metric vs number of images in log scale.}
    \label{fig:ktau}
\end{figure*}

\paragraph{Networks:} We evaluate our algorithm on ResNet18, ResNet50 \cite{he2016deep}, Mobilenetv2 \cite{sandler2018mobilenetv2}, Efficientnet-b0, Efficientnet-lite \cite{tan2019efficientnet}, Mobilenetv3 \cite{howard2019searching}, Deeplabv3-mobilenetv3 \cite{chen2017rethinking}, BERT and ViT model \cite{devlin2018bert}. 

For our experiments, we use the AI Model Efficiency Toolkit (AIMET)$^{1}$ \footnote[1]{$^{1}$AIMET is a product of Qualcomm Innovation Center, Inc., available on GitHub at \url{https://github.com/quic/aimet}}\cite{siddegowda2022neural} to quantize the models to desired bit-widths and use the per-channel weight quantization and symmetric and asymmetric schemes for weight and activation tensors, respectively. We set the quantization range of each quantizer using an MSE based criteria~\cite{nagel2021white}. We do not fix pre-defined bit-widths for any layers or activations in the mixed precision search space and report all mixed precision results in terms of relative BOPs ($r$) with respect to fixed W8A16 representation for each model.

\begin{figure}
    \centering
    \subfloat{\includegraphics[width=0.34 \textwidth]{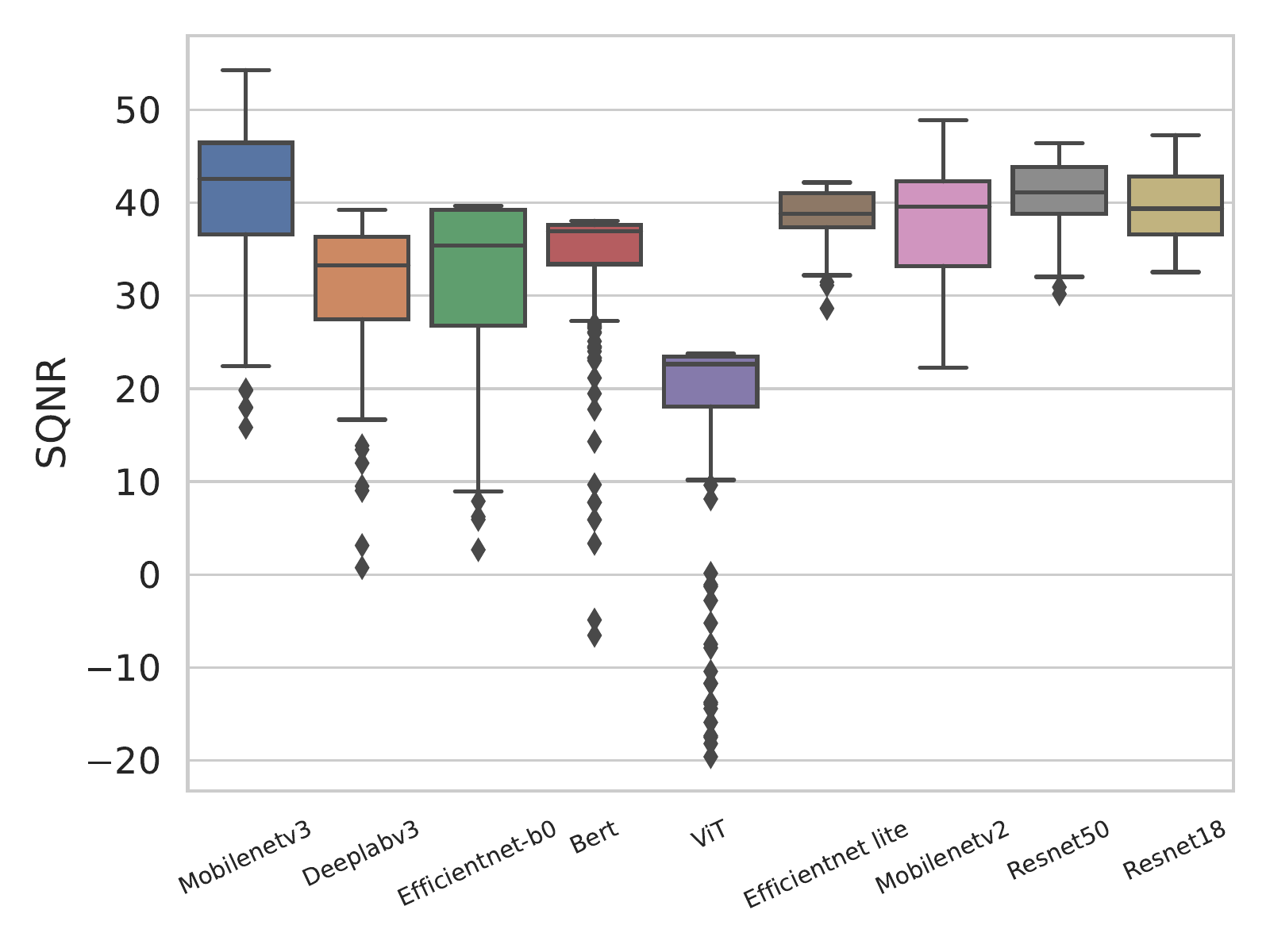}} 
    \caption{Range of SQNR values for W8A8 quantizers in different networks. Networks with layers with outliers or irregular distributions exhibit large SQNR range due to difference in sensitivity of layers to quantization.}
    \label{fig:sqnr-range}
\end{figure}

\subsection{Comparison to Fixed Precision Quantization}
Some networks are very sensitive to quantization due to the presence of layers with large activation ranges with outliers. These ranges are difficult to represent accurately in fixed precision representation. For networks with this issue, mixed precision quantization can be very useful to achieve a better performance-efficiency trade-off by representing problematic layers in high precision while keeping rest of the network in lower bit-widths. 

In Table \ref{tab:fixed-vs-mixed} and \ref{tab:BERT}, we summarize our PTQ mixed precision results using a practical bit-width search space with frequently available bit-widths (W4A8, W8A8 and W8A16) for on-device deployment. We notice that for problematic networks like Mobilenetv3, Deeplabv3, Efficientnet, BERT and ViT, mixed precision quantization outperform its equivalent single precision quantized networks by automatically finding problematic layers and keeping them in higher bits to achieve better performance-efficiency trade-off. 
To understand better why some networks benefit more from mixed precision than others, we visualize the per-layer quantization sensitivity of all networks in Figure \ref{fig:sqnr-range}.
Networks which have layers with outliers like BERT, ViT, Deeplabv3 and Mobilenetv3 exhibit a large SQNR range and have some activations achieving very low SQNR values suggesting significant difference in sensitivity to quantization among layers in these networks. On the contrary, for networks like Resnet18 and Resnet50, we observe a much smaller range of higher SQNR values which explains why mixed precision is less useful for such networks, as many layers have a similar quantization sensitivity. We expect to see in our results that mixed precision works well for the former group, and does not improve significantly on the second group.

\begin{table}
    \centering
    \fontsize{8.25pt}{9.25pt}\selectfont
    \begin{tabular} {  lccccc }
    \toprule
    Task & FP32 & W8A8 ($r$=0.5) &  PTQ MP ($r$=0.5) \\
    \midrule
    RTE & $68.23$ & $49.82$ & $67.14$ \\
    MRPC & $90.38$ & $71.40$ & $86.44$ \\
    SST-2 & $92.43$ & $85.78$ & $91.39$ \\
    STS-B & $88.61$ & $80.63$ & $87.69$ \\
    MNLI & $84.40$ & $74.13$ & $82.97$ \\
    \bottomrule
\end{tabular}
    \caption{MPQ results using W4A8, W8A8, W8A16 bit-width candidates for BERT.}
    \label{tab:BERT}
\end{table}

\begin{table*}
    \centering
    \fontsize{8.25pt}{9.5pt}\selectfont
    \begin{tabular} {  lccccc }
    \toprule
    Model & FP32 & W8A8 AdaRound & MP AdaRound &  W6A8 AdaRound & MP AdaRound\\
    &   &  ($r$=0.50) &  ($r$=0.50) &   ($r$=0.375)  &  ($r$=0.375)\\

    \midrule
    Resnet18 & $69.75\%$ & $69.54\%$ & $69.68\%$ & $69.53\%$ & $69.53\%$\\
    Resnet50 & $76.13\%$ & $75.96\%$ & $75.96\%$ & $75.82\%$ & $75.91\%$\\
    Efficientnet-lite & $75.44\%$ &  $75.34\%$ & $75.36\%$ & $75.23\%$ & $75.18\%$\\
    Efficientnet-b0 & $77.67\%$ &  $14.98\%$ & $76.55\%$ & $12.67\%$ & $69.78\%$\\
    Mobilenetv2 & $71.87\%$ & $70.62\%$ & $70.98\%$ & $70.51\%$ &  $70.56\%$\\
    Mobilenetv3 & $74.04\%$ & $69.96\%$ & $72.83\%$ &$69.76\%$ &  $71.40\%$ \\
    Deeplabv3-Mobilenetv3 & $0.6887$ &  $0.5865$ & $0.6708$ & $0.5827$ & $0.6692$ \\
    \bottomrule
\end{tabular}
    \caption{Comparison between fixed precision AdaRound and MP AdaRounded models for W4A8, W6A8, W8A8, W8A16 search space.}
    \label{tab:vanilla-vs-adarounded}
\end{table*}

Further, to demonstrate generality of our method to lower weight and activation bit-widths, we perform mixed precision analysis on an expanded search space by including 4 and 6 bit quantization for weight and activation tensors. The results are summarized in Table \ref{tab:fixed-vs-mixed_expanded}. Our Mixed precision quantization routine is able to significantly improve accuracy of the networks specially in lower BOPs models by rightly identifying sensitive layers and assigning higher bit-widths to them.



\vspace{-0.05in}

\subsection{Robustness to Calibration Data}
\vspace{-0.05in}

Calibration data plays an important role in most post-training quantization algorithms to estimate quantization range settings for activation feature maps of a network. At times, using an under representative calibration set for PTQ methods can lead to sub-optimum performance, making it challenging to adapt for practical use-cases where little or no task data is available. In this section, we discuss robustness of our algorithm to variation in calibration data used for performing mixed precision, highlighting the wide applicability of our method for real-world scenarios.

\subsubsection{Robustness to variation in images} In order to understand the advantage of using SQNR as a metric to capture per-layer quantization sensitivity, we compare SQNR with other surrogate metrics. 
    An alternate way of capturing the quantization sensitivity of a layer is to measure the degradation in task performance with respect to full precision network upon quantization of that layer. Such a surrogate measure of the relative effect of quantization on different layers works well however it can be computationally expensive due to multiple evaluations on the validation set. Also, the accuracy of such a sensitivity list is affected by how well representative the validation set was of the overall data distribution. To investigate this, we use 5 random subsets of 256 Imagenet images to obtain the quantization sensitivity list using the accuracy and SQNR metric. As we observe in Figure \ref{fig:ktau}, the pareto curve obtained with accuracy metric drastically varies with choice of the subset of images used to creating the sensitivity list. With the same experimental settings, SQNR achieves a much smaller variation in the pareto curve obtained using different subsets of data.

\vspace{-0.1in}
\subsubsection{Robustness to numbers of images}Next, we study the effect of using more images on the accuracy of sensitivity list obtained in Phase 1. To quantify the quality of the sensitivity list obtained using different of number of images, we use the Kendall Tau correlation coefficient between the obtained sensitivity list and the ground truth sensitivity list which we define as sensitivity list obtained using measuring the accuracy degradation on the entire 50K Imagenet validation set as a surrogate measure. Due to the biased nature of the accuracy metric in presence of less or unbalanced calibration data, using accuracy measurements to capture relative sensitivity of layers to quantization can be inaccurate, hence leading to sub-optimal sensitivity list. In similar settings, SQNR being a softer metric, captures the relative sensitivity of layers to quantization better leading to higher Kendall Tau score. 
\par
We also compare SQNR to a per layer FIT~\cite{zandonati2022fit} metric which uses the Fisher information as an efficient approximation to Hessian used by HAWQ~\cite{dong2019hawq} to measure relative sensitivity of layers to quantization. In terms of Ktau score, SQNR performs at par with the FIT metric, without the need of requiring labelled data and backpropagation, making SQNR a great choice for post-training quantization use-cases.

\begin{figure}
    \centering
    \subfloat[]{\includegraphics[width=0.23\textwidth]{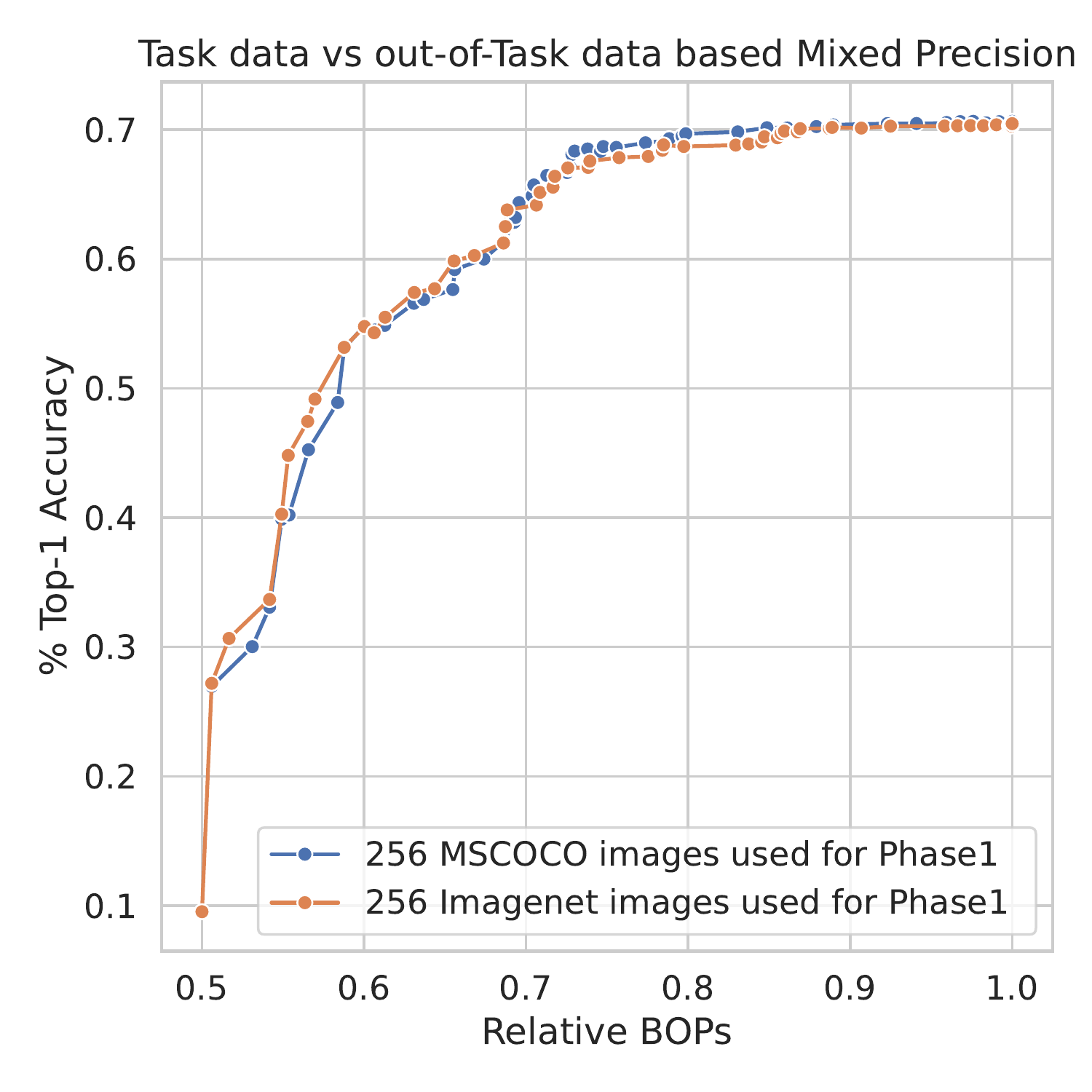}} 
    \subfloat[]{\includegraphics[width=0.23\textwidth]{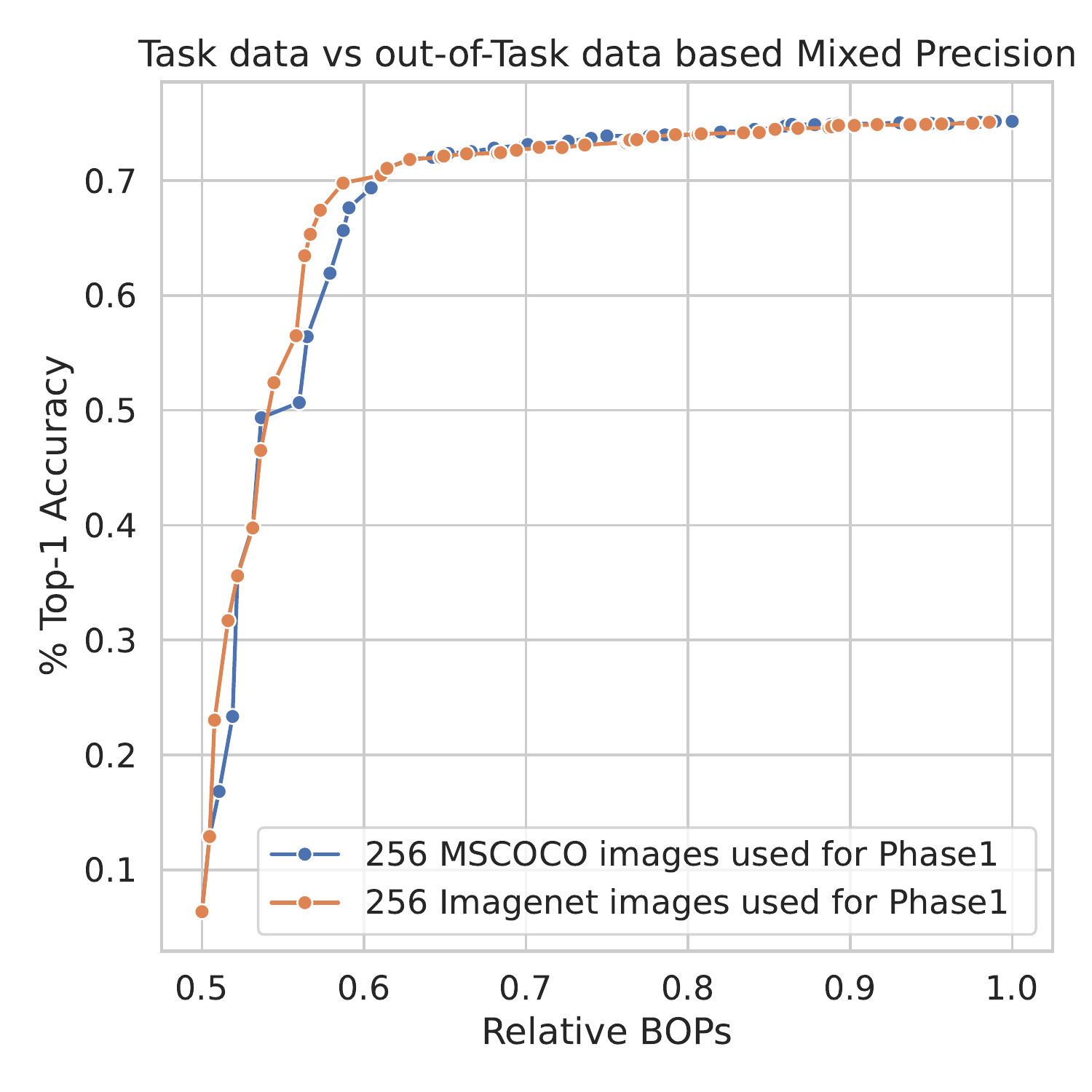}} 

    \caption{Comparison between MP using W4A8, W8A8 bit-width candidates performed using out-of-task data (MS-COCO images) and task-data (Imagenet images) for (a)  Mobilenetv2 (b)  Efficientnet-lite. Both scenarios achieve similar performance-efficiency trade-off.}
    \label{fig:label-free}
\end{figure}

\begin{table*}
    \centering
    \fontsize{8.25pt}{9.0pt}\selectfont
    \begin{tabular} {  lccccc }
    \toprule
    Model & Accuracy Target & Sequential  & Binary  & Binary + Interpolation & Relative BOPs  \\         &   &  Search (hours) &  Search (hours) &   Search (hours)  & $(r)$ \\

    \midrule
    Resnet50 & $75.13\%$ $(-1\%)$ & $12.6$ & $1.6$ & $1.6$ & $0.396$\\
    Resnet50 & $71.13\%$$(-5\%)$ & $14.4$ & $1.6$ & $1.2$ & $0.257$\\
    Efficientnet-lite & $74.44\%$$ (-1\%)$ & $2.3$ & $0.8$ & $0.4$ & $0.859$\\
    Efficientnet-lite & $70.44\%$$ (-5\%)$ & $4.4$ & $0.8$ & $0.5$ & $0.610$\\
    Mobilenetv2  & $70.89\%$$ (-1\%)$ & $5.6$ & $0.4$ & $0.4$ & $0.583$\\
    Mobilenetv2  & $66.89\%$$ (-5\%)$ & $8.5$ & $0.8$ & $0.6$ & $0.365$\\
    Mobilenetv3 & $73.04\%$$ (-1\%)$ & $8.8$ & $0.3$ & $0.3$ & $0.849$\\
    Mobilenetv3 & $69.04\%$$ (-5\%)$ & $14.1$ & $0.8$ & $0.4$ & $0.435$\\
    \bottomrule
    \end{tabular}
    \caption{Run-time comparison between sequential, binary and binary + interpolation based search for W4A8, W8A8, W8A16 MP for performance budgets: $1\%$ and $5\%$ accuracy drop from full precision accuracy.}        \label{tab:phase2-runtime}

\end{table*}

\vspace{-0.15in}

\subsubsection{Robustness to out-of-domain data}
\vspace{-0.05in}

In many practical use-cases, the user may have access to little or no task data due to privacy reasons. As described in the previous sections, labels have no role during Phase 1 of our algorithm. This opens up possibility to use similar task domain data to perform mixed precision analysis in such scenario. To demonstrate this, we use 256 MS-COCO images during both phases of our algorithm for quantization range setting and sensitivity list creation. As summarized in Figure \ref{fig:label-free}, even without any task data (Imagenet images) used, we find the pareto curves obtained in phase 2 for Mobilenetv2 and Efficientnet lite are very similar to pareto curve obtained using Imagenet images in similar experimental settings.

In conclusion, using SQNR to measure the sensitivity of layers to quantization shows robustness to both variation and number of images in the calibration dataset, and also achieves competitive performance with similar out-of-domain data, making it a reliable choice for a hyper-parameter free mixed precision approach for practical use-cases.

\subsection{AdaRound integrated Mixed Precision}
\vspace{-0.05in}

As discussed in the previous section, the performance of low-bit quantization configurations in mixed precision can be improved by integrating AdaRound in our mixed precision routine. Integrating AdaRound in our method has no additional compute overhead other than performing AdaRound on the full precision network for each weight bit-width in the search space. To show improvements with our enhancement method we summarize comparison between fixed precision AdaRound with AdaRound integrated mixed precision in Table \ref{tab:vanilla-vs-adarounded}. Specially for low bit-width ($<$8) quantization, our AdaRound integrated mixed precision performance not only improves over fixed precision AdaRound but also outperforms its fixed precision equivalent even for quantization friendly networks like the Resnet family where mixed precision was not very helpful.

\begin{figure}
    \centering
    \includegraphics[width=0.34\textwidth]{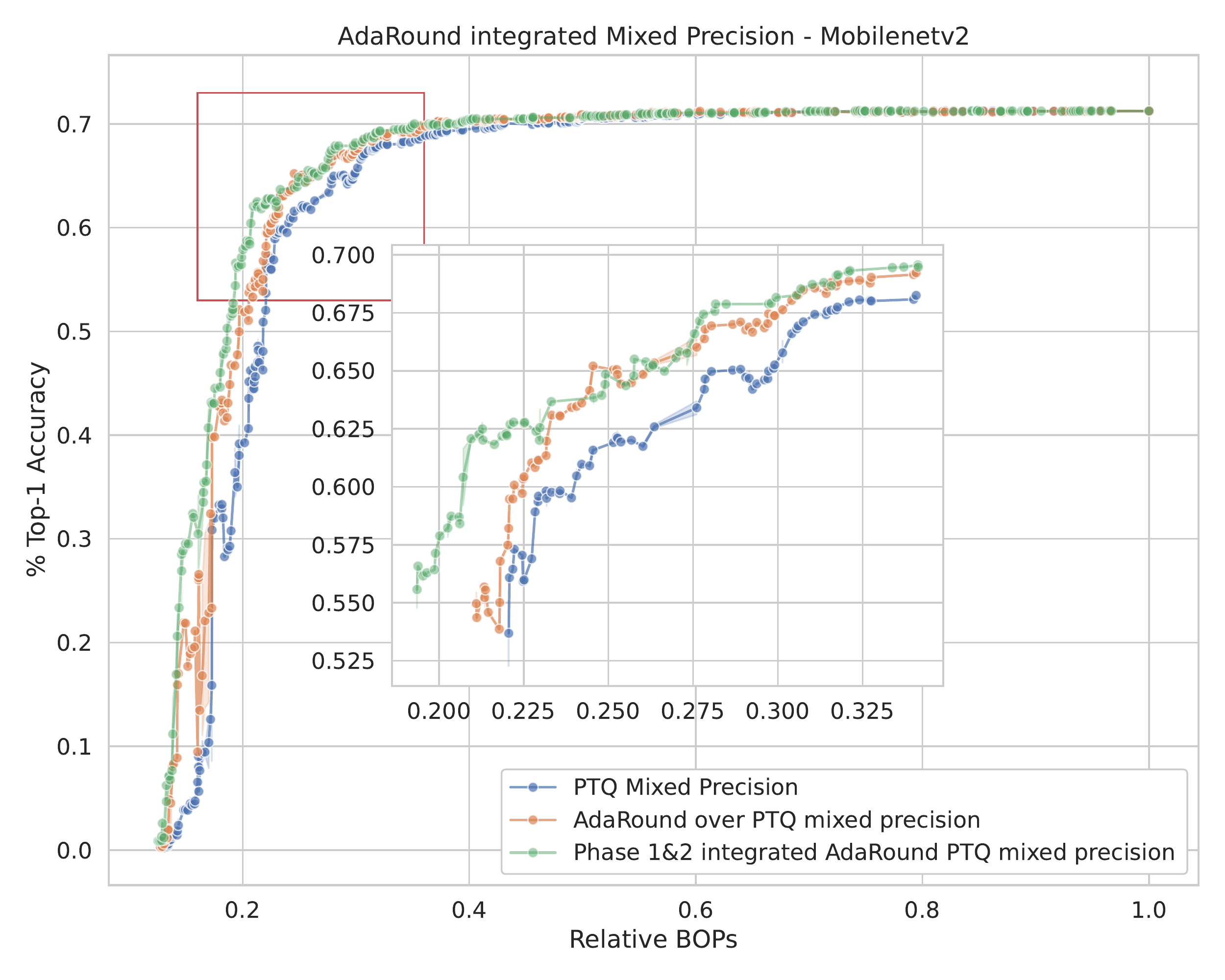}
    \caption{Comparison between PTQ MP, AdaRound over PTQ MP and Phase 1$\&$2 integrated AdaRound PTQ MP.}
    \label{fig:adaround-integration}
\end{figure}

Further, to understand the advantage of interweaving AdaRound in both phases of our mixed precision routine, we conduct ablation on a Mobilenetv2 network with W4A4, W4A6, W6A4, W6A6, W8A6, W6A8, W8A8, W8A16 as our search space and compare the performance of PTQ mixed precision, AdaRound over PTQ mixed precision and AdaRound interweaved in both phases based mixed precision. As we can see in Figure \ref{fig:adaround-integration}, integrating AdaRound in both phases achieves the best performance-effiency trade-off specially in lower BOPs region signifying the importance of correctly capturing the AdaRounded weight-activation quantization trade-off for low bit activation quantization.

\subsection{Phase 2 run-time comparison}
\vspace{-0.05in}

For task performance (accuracy) based target budgets, searching for the least BOPs model satisfying the criteria can be expensive in our trivial sequential search based phase 2. Exploiting the monotonic nature of the pareto curve, we propose to use binary search and its improved variant to improve the overall run-time of our search algorithm. Table \ref{tab:phase2-runtime} shows time taken by these three schemes to find a mixed precision configuration for desirable performance budget of absolute $1\%$ and $5\%$ drop with respect to full precision accuracy.

\vspace{-0.1in}
\section{Conclusions}
\vspace{-0.05in}
In this work we introduced a post-training mixed precision quantization algorithm that sets the mixed-precision bit-widths for practical use-cases. Our algorithm uses little data, requires no hyper-parameter tuning, is robust to data variation, and takes practical hardware considerations into account to automatically select suitable bit-widths for each layer to achieve desirable on-device performance. 
We also discuss integration of our method with complimentary post-training quantization algorithms such as AdaRound, and proposed enhancements to our mixed precision configuration search routine to improve the performance and overall run-time of our algorithm, respectively. 
We show that our post-training mixed precision algorithm finds mixed precision configurations which have a significantly better task performance than their static bit-width equivalents in post-training quantization setting for challenging networks like Mobilenetv3, Deeplabv3, Efficientnet, BERT and ViT.


\bibliographystyle{abbrv}
\bibliography{refs}

\end{document}